\documentclass[letterpaper, 10 pt, conference]{ieeeconf}  
\IEEEoverridecommandlockouts                              
\usepackage{grffile}
\usepackage{color}
\usepackage{graphicx}
\usepackage{cite}
\usepackage{multicol}
\usepackage{subcaption}
\usepackage{algcompatible}
\usepackage{algpseudocode}
\usepackage{siunitx}
\usepackage{amsmath}
\usepackage{amssymb}
\usepackage{array}
\usepackage{caption}
\usepackage[linesnumbered,algoruled,boxed,lined]{algorithm2e}
\usepackage{ifthen}
\usepackage{lipsum,graphicx,subcaption}
\usepackage{comment}
\usepackage{romannum}
\usepackage{float}
\usepackage{url}
\usepackage[usestackEOL]{stackengine}

\graphicspath{ {./Figures/} }
\DeclareMathOperator*{\argmax}{argmax}

\title{\LARGE \bf
Where to go: Agent Guidance with Deep Reinforcement Learning in A City-Scale Online Ride-Hailing Service
}

\author{Jiyao Li$^{1}$ and Vicki H. Allan$^{1}$ 
\thanks{$^{1}$Jiyao Li and  Vicki H. Allan are with the Department of Computer Science, Utah State University, U.S.A.
        {\tt\small (\{jiyao.li, vicki.allan\}@usu.edu)}}%
}

\begin{document}

\maketitle
\thispagestyle{empty}
\pagestyle{empty}

\begin{abstract}
Online ride-hailing services have become a prevalent transportation system across the world. In this paper, we study a challenging problem of how to direct vacant taxis around a city such that supplies and demands can be balanced in online ride-hailing services. We design a new reward scheme that considers multiple performance metrics of online ride-hailing services. We also propose a novel deep reinforcement learning method named Deep-Q-Network with Action Mask (AM-DQN) masking off unnecessary actions in various locations such that agents can learn much faster and more efficiently. We conduct extensive experiments using a city-scale dataset from Chicago.  Several popular heuristic and learning methods are also implemented as baselines for comparison. The results of the experiments show that the AM-DQN attains the best performances of all methods with respect to average failure rate, average waiting time for customers, and average idle search time for vacant taxis.     
\end{abstract}

\section{INTRODUCTION}
Online ride-hailing services such as Uber \cite{uber} and Lyft \cite{lyft} have significantly changed the way of people travel \cite{sadowsky2017impact}. Instead of hailing taxis on streets like traditional taxi services, customers send service orders through mobile apps. Such online services can optimize the transportation resources such that traffic congestion is greatly alleviated and gas emissions are substantially reduced \cite{rodier2018effects}.

One of the key challenges of the online ride-hailing services is to direct vacant taxis such that the balance between supply and demand can be met. In some big cities, even though there are plenty of available taxis, a large number of customers still fail to be served because the available taxis are not close enough to reach the passengers in a timely manner. If vacant taxis could be guided to the areas of current and future expected need, it would dramatically increase the number of passengers who can be served, reduce passengers' waiting time and taxis' idle search time.

There are several challenges in scheduling supplies (guiding taxis) to meet demands: $(\romannum{1})$ the demand in one location of a city fluctuates constantly such that it is extremely hard to make an accurate prediction; $(\romannum{2})$ customers cancel their requests and turn to alternative services if their waiting time exceeds their patience duration. $(\romannum{3})$ effort is wasted if too many vacant taxis move to an area of need. 

Most of the current research papers that study the taxi repositioning problem have the following limitations: $(\romannum{1})$ Most existing studies use private datasets and assume that each zone is a polygon with regular dimension, but in reality, public datasets like Chicago and New York City, have irregularly shaped areas. $(\romannum{2})$ much current work assumes that passengers wait indefinitely until they are picked up by taxis. However in most cases, passengers' patience is so limited that they would change to another service after a certain period of waiting. 

To address the challenges and the limitations of the online ride-hailing service, we propose a novel deep reinforcement learning solution known as \textbf{D}eep-\textbf{Q}-\textbf{N}etwork with \textbf{A}ction \textbf{M}ask (AM-DQN). The main contributions of the paper are described below:
\begin{itemize}
    \item We design a new reward scheme that combines the concerns of the waiting time of passengers and idle search time of drivers.
    \item To address dynamic action space, we propose a new Deep-Q-Network named AM-DQN using an action mask for the various dimension of action space such that agents can learn much faster and more efficiently.
    \item We use a city-scale dataset (1,185,018 order records) and implement other popular heuristic and learning methods as baselines to verify the effectiveness of our solution.
\end{itemize}

The organization of the rest paper is as follows: the recent related studies of online ride-hailing services and reinforcement learning are discussed in Section \ref{s2}. In Section \ref{s3}, we introduce several definitions and their notation. The Taxis Repositioning Problem (TRP) is defined. In Section \ref{s4}, we propose a new reward scheme and a novel deep reinforcement learning method named AM-DQN for the online ride-hailing service. A city-scale dataset and several popular baselines are used to verify the effectiveness of our method in Section \ref{s5}. Section \ref{s6} is the conclusion of our achievements and contributions.

\section{Related Work}
\label{s2}
\textbf{Online Ride-hailing Services.} \cite{wen2017rebalancing} adopts a Deep-Q-Network to guide vacant taxis to desired places. Extensive experiments show reduction in waiting time and relocation cost, but the authors assume that customers wait indefinitely for the service ignoring the patient duration of passengers. \cite{lin2018efficient} proposes a multiagent reinforcement learning method to coordinate vehicles locating between two adjacent zones. The solution can improve the performance of the system but the cooperation between vehicles still requires hard coded rules rather than being totally learned. \cite{li2020balancing} uses a hybrid solution to direct vacant taxis around the city such that the balance between supplies and demands can be kept, but the greedy idle movement strategy lacks flexibility and is difficult to adapt to the dynamic environment. Based on the work of \cite{li2020balancing}, \cite{icaart22} proposes the lottery scheme in ride-matching and applies Q-learning in vehicle relocation instead of greedy idle movement, but the tabular method is limited when the dimension of state gets to be larger. \cite{de2020efficient} applies a Monotonic Value Factorization method to let all available vehicles cooperate with each other for the dispatching, but the method uses large amounts of computational resources such that it is only suitable for small scale datasets. \cite{zheng2022supply} proposes an action sampling policy with DQN to schedule taxis to neighbors or global popular locations.  They use a public dataset of New York City to conduct extensive experiments. However they do not consider the dynamic property of action space at each location.  The learning efficiency is affected by the unnecessary actions. \cite{qu2014cost} introduces a novel recursive approach to find the optimal relocation route for vacant taxis. The time complexity is high although it can increase drivers' profit. \cite{li2019ride} proposes a polar coordinate method to direct vehicles to find nearby passengers. It can save riders' waiting time and drivers' traveling cost, but it fails to find the optimal rider for a driver. \cite{ma2022efficient} and \cite{li2021heterogeneous} present a novel policy gradient approach with the attention mechanism to solve the pickup and dropoff problem, both articles only consider single vehicle scenario rather than multiple vehicles.

\textbf{Reinforcement Learning.} \cite{watkins1992q} proposes a model free learning solution named Q-learning based on the TD error learning \cite{sutton2018reinforcement}. \cite{mnih2013playing} proposes a Deep-Q-Network (DQN) that performs playing Atari 2600 games. Based on the work of \cite{mnih2013playing}, \cite{mnih2015human} uses a target network to stabilize the learning process. The authors in \cite{tampuu2017multiagent} introduce a cooperative DQN by modifying the reward scheme.  However, they only use one Atari game (Pong) to demonstrate the effectiveness of the solution. \cite{rashid2018qmix} proposes a novel multiagent learning method named QMIX to let agents learn to cooperate with each other to finish a specific task.

\section{Problem Statement}
\label{s3}
\subsection{Definitions and Notation}
\textbf{DEFINITION 1: Zone.} A city is divided into zones $Z=\{1,2,...,z,...,|Z|\}$, where $Z$ is the set of zones and $z$ is a variable representing zone ID. The set of zones partition the city into disjoint areas. 

\begin{figure}[h]
    \centering
    \begin{subfigure}{0.2\textwidth}
        \centering
        \includegraphics[width=0.5\textwidth]{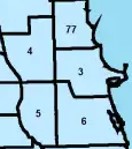}
        \caption{Zones} 
        \label{fig:zones}
    \end{subfigure}
    \begin{subfigure}{0.2\textwidth}
        \centering
        \includegraphics[width=0.7\textwidth]{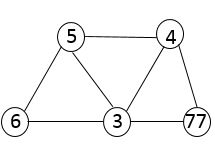}
        \caption{Zone Network} 
        \label{fig:network}
    \end{subfigure}
\caption{Illustration of a set of zones and its zone network. (a) A portion of zones in a city. (b) A zone network is an undirected graph, where each node represents a zone in (a), and edges denote adjacency.} 
\label{fig:zone_network}
\end{figure}

\textbf{DEFINITION 2: Zone Network.} A zone network $G$ is denoted by an undirected graph $G=(Z,E)$, where $Z$ is the set of zones of the city and $edge(z_i,z_j) \in E$ represents connectivity between adjacent zones, as shown in Fig.\ref{fig:zone_network}.

\textbf{DEFINITION 3: Order.} A customer order $o$ is denoted by a tuple $o=({id}_o, t_o, s_o, d_o, p_o, t_d)$, where ${id}_o$ is an unique identification number of the order $o$, $t_o$ is the time stamp when customers send the order $o$, $s_o \in Z$ and $d_o \in Z$ are the source and destination, $p_o$ is the patience duration during which the customer  will not cancel the request, and $t_d$ is the estimated trip duration between pickup and dropoff.

We use $O=\{o_1, o_2,...,o_{|O|}\}$ to denote all the orders. Note that there are two final states for each order: $(\romannum{1})$ an order $o$ can be picked up during the patience period $\left[t_o, t_o+p_o \right]$ at source $s_o$, and dropped off at $d_o$ after $t_d$. In this state, the order $o \in O^+$, where $O^+$ is the set of orders that can be served successfully; $(\romannum{2})$ an order $o$  may fail to be picked up during its patience duration. In this state, the order $o \in O^-$, where $O^-$ is the set of orders that failed to be served. In terms of passengers' satisfaction, two important performance metrics are described in the following:
\begin{itemize}
    \item \textbf{Failure Rate $\mathcal{FR}$.} The failure rate denotes what percentage of customers fail to be served, which is defined by $\mathcal{FR} = \dfrac{|O^-|}{|O|}$.
    \item \textbf{Waiting Time $t^o_w$.} The waiting time of an order $o$ is the time duration between the moment when the customer sends the request and the moment when the customer is picked up.
\end{itemize}

\textbf{DEFINITION 4: Driver.} A driver $d$ is an agent who drives a taxi.  The driver is denoted by a tuple $d=(id_d, t, z_d, s_d)$, where $id_d$ is an unique identification number of the driver, $t$ is current time stamp, $z_d$ is zone ID where the driver is currently located, and $s_d$ is a boolean variable indicating if the driver is busy serving customers.

We use $D=\{d_1, d_2,..., d_{|D|}\}$ to denote all drivers of a city. Note that there are two states of drivers based on $s_d$: $(\romannum{1})$ if $s_d=true$, it indicates that the driver is unavailable as it is currently serving riders; otherwise $(\romannum{2})$ the driver has no work and it is available for serving customers.The following metric captures the drivers costs:
\begin{itemize}
    \item \textbf{Idle Search Time $\mathcal{IST}_d$.} The Idle Search Time $\mathcal{IST}_d$ of driver $d$ denotes the amount of time during which a taxi is recognized as vacant ($s_d$ is false) until it picks up an order.
\end{itemize}

\subsection{Taxis Repositioning Problem and Unified Objective Function}
\textbf{DEFINITION 5: Taxis Repositioning Problem (TRP).} Given a zone network of a city $G$, a set of drivers $D$, a set of orders $O$ that arrive over time, the TRP is to relocate idle drivers ($s_d$ is false), such that the failure rate and waiting time of orders as well as idle search time of drivers can be minimized.

\section{Methodology}
\label{s4}
\subsection{Markov Decision Process (MDP)}
\label{s4:mdp}
Typically, the Taxis Repositioning Problem (TRP) is seen as a sequential decision-making problem that can be modeled as a Markov Decision Process (MDP). Each available driver is considered as an agent and each agent shares the same policy scheme. Agents interact with the environment depending on current state and the reward signal. The goal of each agent is to maximize its reward gain for a long term $\sum \limits_{t=T_0}^T \gamma^t \cdot r_t$, where $T_0$ is the current time stamp, $r_t$ is the reward at time stamp $t$ and $\gamma$ is a discount factor. The common objective of MDP to learn the action-value function $Q(s,a)$ (the expected return taking action $a$ at state $s$) with an optimal policy $\pi$, $Q(s,a) = \mathbb{E}_\pi [\sum \limits_{t=T_0}^T \gamma^t \cdot r_t | s_t=s, a_t=a]$, where $s_t$ is the state and $a_t$ is the action at time stamp $t$.

The key components of MDP are described as follows.

\textbf{State.} The state $s_t$ is defined as a three-tuple which denotes temporal and spatio status. Formally, we define $s_t=(t,da,z) \in S$. In the tuple, $t$ is the current time stamp; $da$ is the day of the week. We consider $da$ as an element of the state tuple because the order pattern varies from day to day within a week. $z \in Z$ is the current zone ID of the driver. 

\textbf{Action.} The action $a_t$ is selected by each available driver to determine the route of a taxi at time $t$. Each available driver can either move to one of the adjacent zones or stay in the current zone. Note that the action space at various zones changes as the degree of each zone (the number of edges incident to $z$) $deg(z) = |edge(z,adj\small(z\small))|$, where $adj(z)$ is the set of adjacent nodes of $z$. We define the action space as a dynamic set $\mathcal{A}(z)=\{k\}^{|deg(z)+1|}_{k=1}$ indicating that drivers at zone $z$ can move to one of $|deg(z)|$ adjacent zones or stay in zone $z$.

\textbf{Reward.} The reward $r_t$ is an immediate feedback for agents after they execute actions at time $t$. The goal of agents is to learn an optimal policy $\pi$ such that the failure rate, the waiting time of orders, and idle search time of drivers can be minimized over the long term (other words, maximize the reward over the long term). Therefore, the reward scheme is supposed to be designed for helping agents learn the optimal policy $\pi$ that can achieve better performance. In our reward scheme shown in Algorithm ($\ref{algo:reward}$), an agent gets a positive reward if it is assigned with an order $o$, the scale of the reward is determined by customer's patient duration $p_o$ and waiting time $t^o_w$. The purpose of the scale setting is to motivate drivers to pick up passengers as early as possible, preventing passenger loss and lessening the waiting time $t^o_w$ of order $o$. An agent gets a zero reward if it fails to be assigned to an order. In the simulated experiment, $C_r$ is set to 1.

\begin{algorithm}
    \SetKwInOut{Input}{Input}
    \SetKwInOut{Output}{Output}
    \Input{Driver $d$ and his/her serving order $o$.}
    \Output{Reward value at time stamp $t$.}
    \eIf{Driver is working on current order $o$ }
     {
        $r_t = (p_o - t^o_w) \cdot C_r$\\
     }
     {
        $r_t=0$\\
     }
     \Return $r_t$
    \caption{Reward Scheme}
    \label{algo:reward}
\end{algorithm}

\textbf{State Transition.} The state transition is from the current state $s_t$ to another state after agents execute action $a_t$. In our MDP, there are two transition types: $(\romannum{1})$ when an agent fails to be assigned with order after executing action $a_t$, the current state $s_t=(t,da,z)$ will become  $s_{t+1} = (t+1, da, o(z))$, where $o(z)$ indicates either zone $z$ or one of adjacent zones of $z$; $(\romannum{2})$ when an agent is assigned with an order $o$ after executing action $a_t$, the current state $s_t=(t,da,z)$ will become $s_{t+t_d^o} = (t+t_d^o, da, d_o)$ state where $t_d^o$ is the trip duration and $d_o$ is the destination zone of $o$. 

\textbf{Discount Factor.} The discount factor $\gamma$ determines the impact of future rewards. $\gamma \in [0,1]$.

\subsection{DQN with Action Mask (AM-DQN)}
\label{subsec:am-dqn}
As previously mentioned, the dimension of the action space at each zone is not fixed because the degree of zones ($deg(z)$) in the zone network $G$ is typically different. Normally, the maximized dimension of action space is applied for the Deep-Q-Network \cite{mnih2015human} to solve the problem\footnote{In the paper, the authors set the action space dimension as 5 across all Atari 2600 games; however, most games, like Pong, Breakout, only use 2 or 3 actions for control.}. Although such a solution is feasible, it is inefficient because many unnecessary actions have to be learned. For example, as shown in Fig. \ref{fig:network}, the maximized dimension of action space is 5 (the degree of zone 3 is 4), while in zone 6, its necessary dimension of action space is only 3 (the degree of zone 6 is 2).

\begin{figure}[H]
    \centering
    \includegraphics[width=0.45\textwidth]{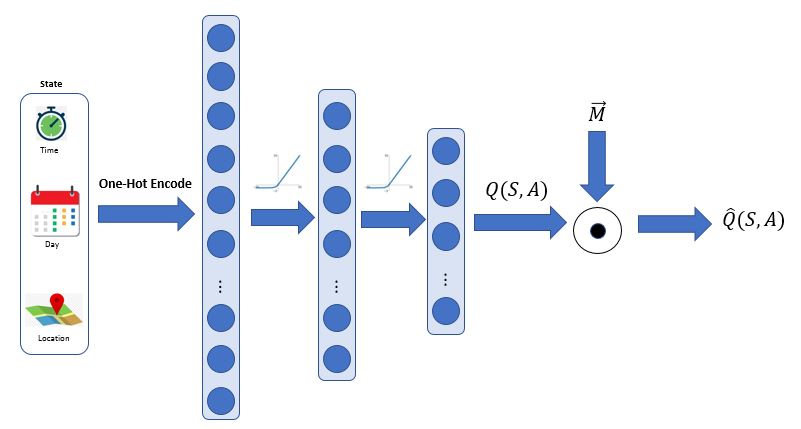}
    \caption{AM-DQN Architecture}
    \label{fig:am_dqn}
\end{figure}

To solve the dynamic action space problem, a method named \textbf{DQN} with \textbf{A}ction \textbf{M}ask (AM-DQN) is proposed to only learn necessary actions at each zone. The general idea is shown in Fig. \ref{fig:am_dqn}: the state is input into the AM-DQN through one-hot encoding; the output of the AM-DQN $Q(S,A)$ is a vector whose length is the maximized dimension of action spaces $|Q(S,A)|=\Delta(G)+1$, where $\Delta(G)=\max \limits_z deg(z)$; the action mask $\overrightarrow{M}$ is used to filter out the unnecessary actions in $Q(S,A)$ by element-wise product with $\overrightarrow{M}$ and gets the new action function $\hat{Q}(S,A) = Q(S,A) \odot \overrightarrow{M}$. An appropriate action of an agent can be selected according to $\hat{Q}(S,A)$.

\begin{figure}[H]
    \centering
    \includegraphics[width=0.45\textwidth]{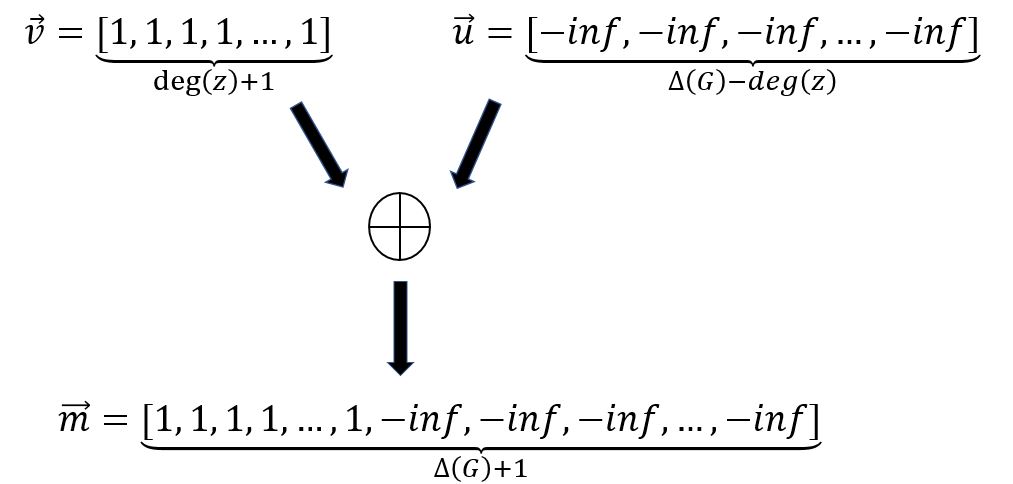}
    \caption{Illustration of action mask generation}
    \label{fig:mask_gen}
\end{figure}

\textbf{Action Mask.} The action mask $\overrightarrow{M}$ is a vector that has the same length as $Q(S,A)$, which is $\Delta(G)+1$. The generation of action mask is described in Fig. \ref{fig:mask_gen}.First, two vectors named $\overrightarrow{u}$ and $\overrightarrow{v}$ are created, $\overrightarrow{u}$ is filled with 's and its length is also the number of necessary actions $\mathcal{A}(z)$ in a zone $z$, $\overrightarrow{v}$ is filled with $-inf$ and its length is the number of unnecessary actions in a zone $z$. The vectors $\overrightarrow{u}$ and $\overrightarrow{v}$ are concatenated  as $\overrightarrow{M}$ of length of $\Delta(G)+1$. We use the action mask $\overrightarrow{M}$ to multiply $Q(S,A)$ in element-wise way, such that elements in the unnecessary action part of $Q(S,A)$ turns out to be $-inf$, so the unnecessary actions of the zone are masked off and cannot be selected.

\textbf{Action Selection.} A simulated annealing policy is used for action selection. The general idea is that random actions are preferred to be selected at the early stage.  As the network matures, the policy tends to select actions according to the output of the network. The details of the algorithm is described in Algorithm (\ref{algo:act}), where $\epsilon=0.1$ and $\lambda=20$.  

\begin{algorithm}
    \SetKwInOut{Input}{Input}
    \SetKwInOut{Output}{Output}
    \Input{Available driver $d$ and current time stamp $t$}
    \Output{The selected action $a_t$}
    $\Theta$ = $\epsilon + (1-\epsilon) \cdot e^{-\frac{t}{\lambda}}$ \\
    Generate a random number $\mathcal{N}$ \\
    \eIf{$\mathcal{N} > \Theta$}
     {
        $a_t = \argmax \limits_{A \in \mathcal{A}(G)} \big( Q(S,A) \odot \overrightarrow{M} \big)$\\  
     }
     {
        Randomly select an action $a_t$ from $\mathcal{A}(z_d)$
     }
     \Return $a_t$
    
    \caption{Simulated annealing policy}
    \label{algo:act}
\end{algorithm}

\textbf{AM-DQN Training.} For training purposes, two identical network architectures are used: a policy network which updates weights and biases each iteration for learning the action value function $Q(S,A)$; a target network which is synchronized with the policy network periodically is adopted to compute the action value of the next state. The policy network is trained by Stochastic Gradient Descent (SGD) \cite{robbins1951stochastic} and Backpropagation \cite{rumelhart1986learning} to minimize the mean squared loss $\mathcal{L}$ in Eq.(\ref{eq:loss}).
\begin{equation}
    \mathcal{L} = \mathbb{E} \left( Q(s_t^i, a_t^i; \theta) - ( r_t^i + \gamma \max_{a'} \left( Q(s_{t'}^i, a'; \theta') \odot \overrightarrow{M}) \right) \right) ^2
\label{eq:loss}
\end{equation}

In the Eq.(\ref{eq:loss}), $\theta$ and $\theta'$ are the parameters (weights and biases) of the policy network and the target network separately. $r_t^i + \gamma \max_{a'} \left( Q(s_{t'}^i, a'; \theta') \odot \overrightarrow{M} \right)$ is the target value with the parameters $\theta'$ that are used for stability purposes. 

The AM-DQN algorithm is presented in Algorithm (\ref{algo:dqn}). At first, we initialize a queue $\mathcal{R}$ in length of $l_{rb}$ as a replay buffer, and we create the policy network and target network with random parameters $\theta$ and $\theta'$ where $\theta' = \theta$ initially. Then our model is trained within a specific maximum timestep through a simulator. In each timestep, each driver that is available interacts with the simulator, stores his/her experience as a transition into $\mathcal{R}$ and computes the loss function $\mathcal{L}$ by sampling transitions from $\mathcal{R}$. After that, Stochastic Gradient Descent (SGD) is performed and Backpropagation is used to update parameters in each layer of the policy network. Moreover, the target network is synchronized with the policy network periodically.   

\begin{algorithm}
    Initialize a queue $\mathcal{R}$ as replay buffer of which capacity if $l_{rm}$.\\
    Create a policy network with random weights and biases $\theta$.\\
    Create a target network of which weights and biases $\theta' = \theta$.\\ 
    \For {$t$ = 1 to $T_{max}$} {
        \For {each $d$ in $D$} {
            \If{$d_s$ is false}{
                Generate an action $a_t$ from Simulated Annealing Policy in state $s_t$.\\
                Execute action $a_t$ in simulator and observe reward $r_t$ and next state $s_{t'}$.\\
                Store the transition $(s_t, a_t, r_t, s_{t'})$ into $\mathcal{R}$.\\
                Sample $N_{mini}$ transitions from $\mathcal{R}$ randomly.\\
                $\mathcal{L} = 0$\\
                \For {i = 1 to $N_{mini}$}{
                    $ y_i = r_i + \gamma \max_{a'} \left( Q(s_{t'}, a'; \theta') \odot \overrightarrow{M} \right) $\\
                    $\mathcal{L} = \mathcal{L} + \left( y_i - Q(s_t, a_t; \theta) \right)^2$ \\
                }
             }
        }
        Perform the SGD on $\mathcal{L}$ with respect to $\theta$ and update it
        $\theta = \theta + \alpha \bigtriangledown_{\theta} \mathcal{L}$\\
        \If{t mod $C_{sync} == 0$}{
            Synchronize the target network $\theta' = \theta$\\
        }
    }
    \caption{AM-DQN Training}
    \label{algo:dqn}
\end{algorithm}

\section{Experiment}
\label{s5}
To evaluate the effectiveness of the AM-DQN, we conduct experiments and compare our model with several other popular methods using a large-scale dataset. The AM-DQN and all other baseline solutions are implemented in Python 3.9.7 and Pytorch 3.9 with CUDA 10.2 $\&$ CUDNN 7.0 . All the methods are run on the machine with GeForce RTX 2080 8G GPU and Intel Core i7-7700K CPU with 64GB memory.

\subsection{Experimental Settings}
\label{subsec:setting}
We use a city-scale dataset from the City of Chicago \cite{city2018chicago}. The dataset includes 1,185,018 taxi trip orders in August 2019. Each taxi trip order is composed of occurrence time of an order, source zone index, destination zone index, trip duration time and patient period. In our simulated experiment, the whole episode is 44640 cycles (one month), and the interval of a simulated cycle is 1 minute. A fixed number of taxis are arranged at the beginning of the experiment. Specifically, there are 10 taxis at each zone, a total of 770 taxis available for the online ride-hailing service.

For the AM-DQN model parameters, the policy network and the target network are built with three fully connected layers respectfully, each of which has 1524 input units, 256 hidden units 10 output units and eLU activation function is applied. We set the capacity of the replay memory $l_{rm}=10,000$, the mini-batch of sampling is $N_{mini} = 32$, the target network updated period $C_{sync} = 10000$, the discount factor $\gamma=0.99$ and the learning rate $\alpha=0.000001$. 

\subsection{Performance comparison}
To verify the effectiveness of the AM-DQN, we implement several popular baseline methods for performance comparison. The brief descriptions of each method are described as following:
\begin{itemize}
    \item \textbf{Random.} The agent moves to one of adjacent zones or stays in current zone randomly.
    \item \textbf{Greedy.} The agent moves to one of adjacent zones (including current zone) with highest demand.
    \item \textbf{Demand Based.} The difference between the amount of demand and supply at adjacent zones and current zone is computed, and agent will select a zone with a probability in proportion to the difference.
    \item \textbf{Q-Learning\cite{watkins1992q}.} The agent interacts with the environment and stores the experience in a table. The reward scheme in \cite{icaart22} is used for the training and $\epsilon$-greedy policy is applied for the action selection and $\epsilon = 0.1$.
    \item \textbf{AM-DQN.} The Deep-Q-Network with Action Mask we propose in Section \ref{s4}.  
\end{itemize}

\begin{figure}[H]
    \includegraphics[width=0.55\textwidth]{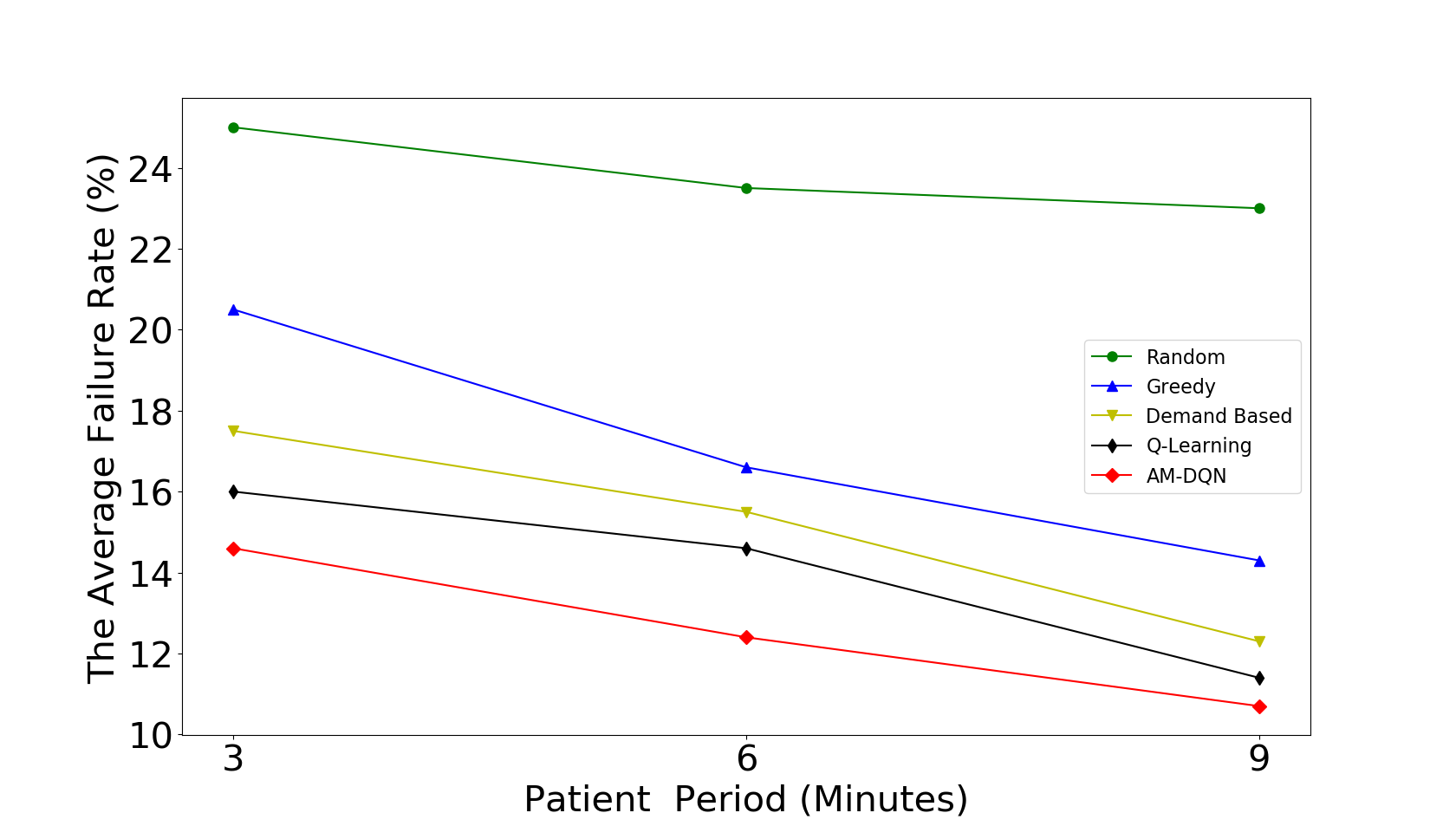}
    \caption{The Average Failure Rate}
    \label{fig:failure_rate}
\end{figure}

The comparison results of the average failure rate along with various patient period are shown in Fig. \ref{fig:failure_rate}. We observe that more customers can be served if they have a longer patient period. Among all the compared methods, the AM-DQN achieves the lowest failure rate of all the methods. This is mainly because the AM-DQN only focuses on learning the necessary actions at each state, ignoring the unnecessary actions. We also observe that the Q-Learning method fluctuates slightly because the tabular method has difficulty converging with states of high dimension.

\begin{figure}[H]
    \centering
    \includegraphics[width=0.55\textwidth]{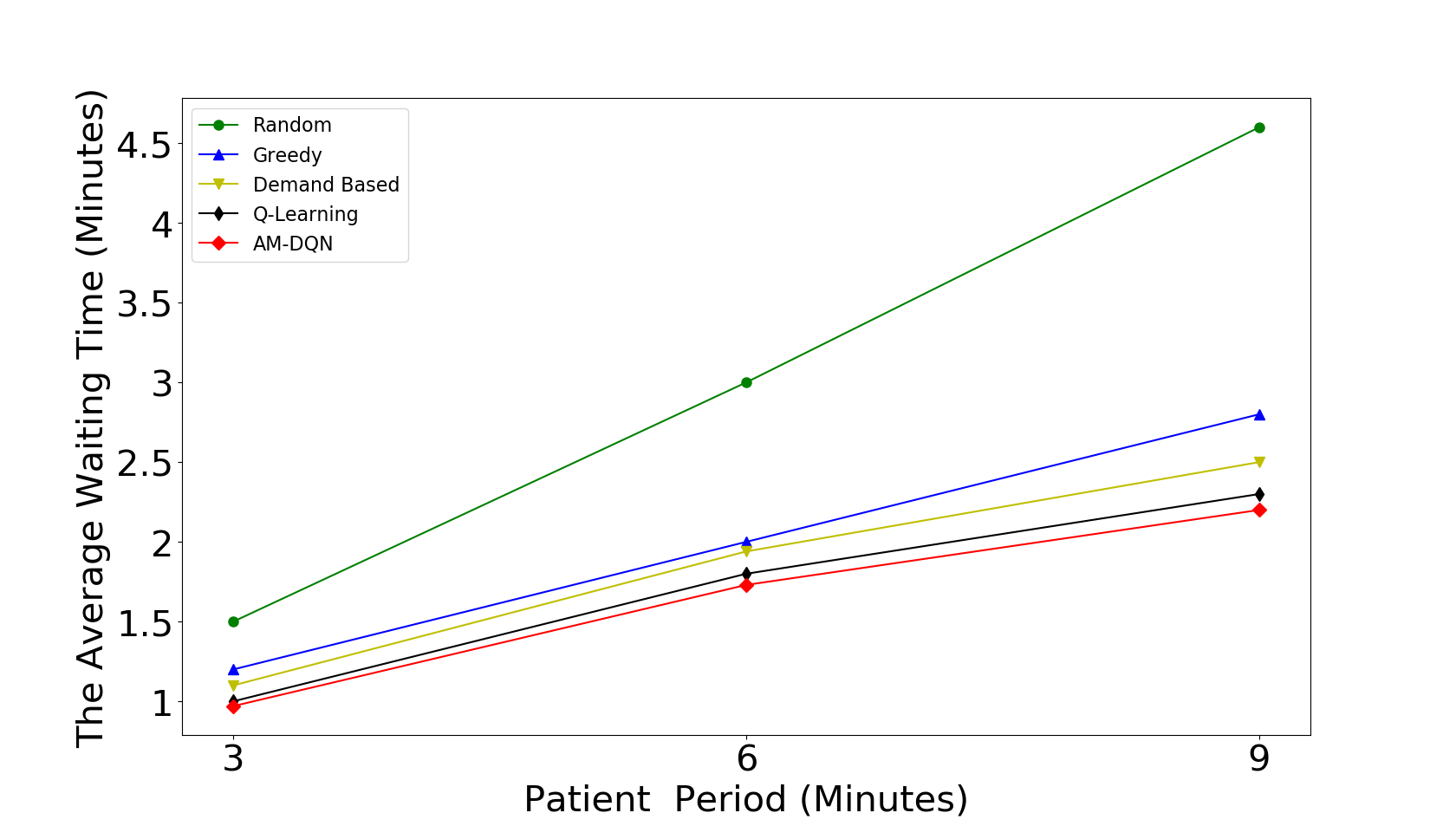}
    \caption{The Average Waiting Time}
    \label{fig:wait_time}
\end{figure}

The Fig. \ref{fig:wait_time} shows the comparison results of the average waiting time of passengers with different patient period. We see that the average waiting time gets to be longer when customers have more patience for the service. We also see that the average waiting time of AM-DQN is the lowest of all solutions. This occurs because the reward scheme of the AM-DQN considers waiting time as one of the reward factors to motivate drivers to pick up passengers as soon as possible. We also observe that the Q-Learning is better than the Demand Based since it considers the reward for a long term period while Demand Based is myopic, on the other hand, Demand Based is better than Greedy because it considers the spatial distribution of both orders and taxis.     

\begin{figure}[H]
    \centering
    \includegraphics[width=0.55\textwidth]{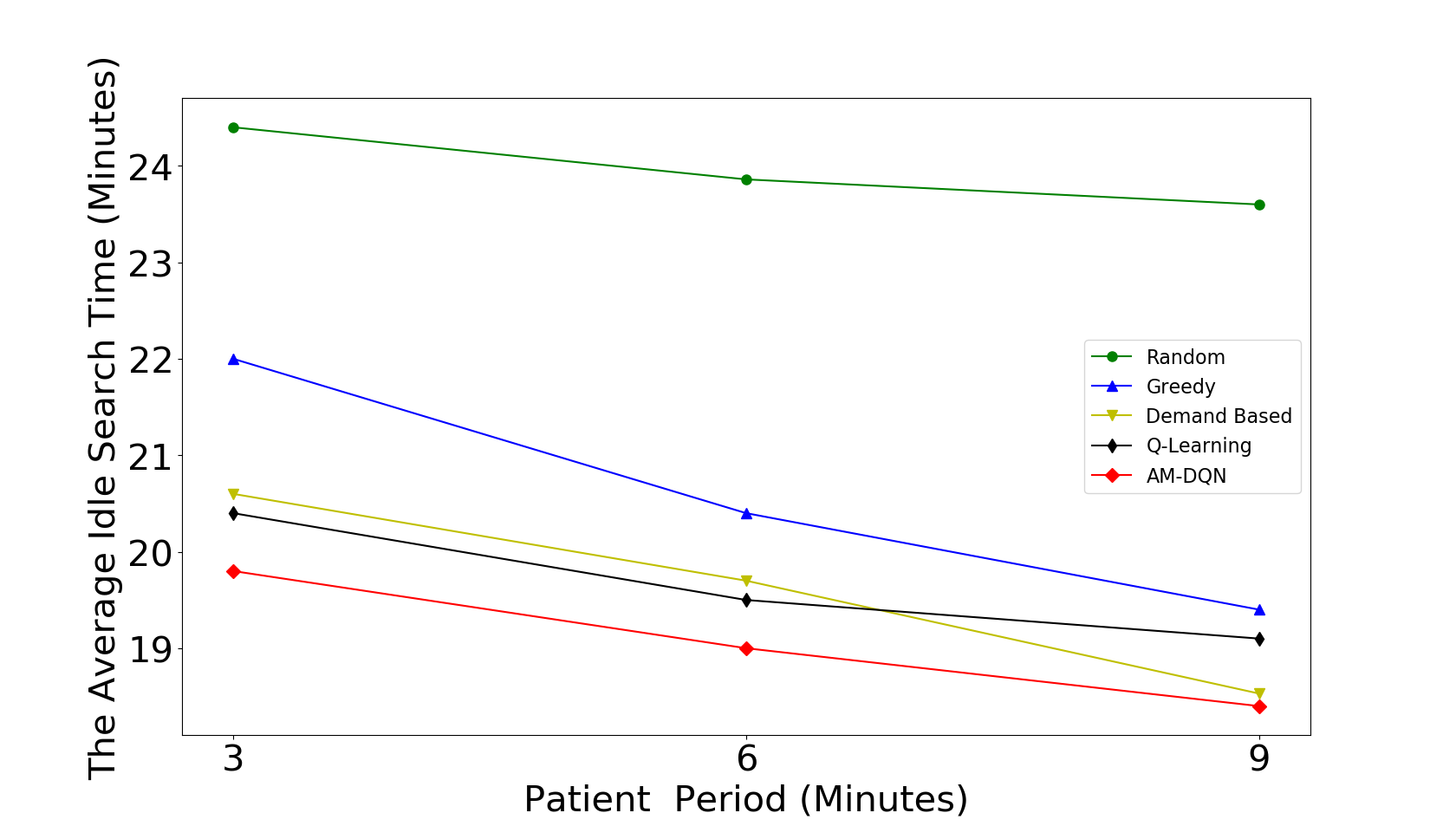}
    \caption{The Average Idle Search Time}
    \label{fig:idle_time}
\end{figure}

As shown in Fig. \ref{fig:idle_time}, we observe that the average idle search time of the AM-DQN is better than other methods when the patient period is short. This is because the reward of the AM-DQN makes agents more sensitive where to pick customers especially when they easily give up requesting the service.

\section{Conclusion}
\label{s6}
This work has three contributions. First, we design a new reward scheme that is time and cost sensitive, balancing the benefits between passengers and drivers. Secondly, we propose a novel deep reinforcement learning method named AM-DQN that can adapt to the variation of the action space, the proposed solution can manage the balance between supplies (taxis) and demands (orders) around a city. Finally, we conduct extensive experiments with a city-scale dataset of Chicago, the experiments show that our solution is the best of all methods in terms of the average failure rate and the average waiting time of customers, as well as the average idle search time of drivers.

\bibliographystyle{ieeetr}
\bibliography{main}

\end{document}